\documentclass{article} 
\usepackage{iclr2024_conference,times}

\usepackage[utf8]{inputenc} 
\usepackage[T1]{fontenc}    
\usepackage{url}            
\usepackage{booktabs}       
\usepackage{nicefrac}       
\usepackage{microtype}      
\usepackage{graphics,graphicx,color}

\usepackage{algorithm}
\usepackage{algorithmic}
\usepackage{enumitem}

\usepackage{amsfonts}       
\usepackage{amsmath}       
\usepackage{amssymb}

\newcommand{\figref}[1]{Fig.~\ref{#1}}    

\newcommand{\tabref}[1]{Table~\ref{#1}}

\usepackage{xspace}
\makeatletter
\DeclareRobustCommand\onedot{\futurelet\@let@token\@onedot}
\def\@onedot{\ifx\@let@token.\else.\null\fi\xspace}
\makeatother

\usepackage{xr-hyper}
\makeatletter
\newcommand*{\addFileDependency}[1]{
  \typeout{(#1)}
  \@addtofilelist{#1}
  \IfFileExists{#1}{}{\typeout{No file #1.}}
}
\makeatother

\usepackage{xcolor}

\definecolor{ourorange}{HTML}{e19c24}
\definecolor{ourgreen}{HTML}{97b567}
\definecolor{ourred}{HTML}{ec6235}
\definecolor{ourblue}{HTML}{5e81b5}
\definecolor{ourgrey}{HTML}{919191}

\usepackage{subcaption}

\DeclareMathAlphabet{\mathsfit}{\encodingdefault}{\sfdefault}{m}{sl}
\SetMathAlphabet{\mathsfit}{bold}{\encodingdefault}{\sfdefault}{bx}{n}

\def\gA{{\mathcal{A}}}

\def\gS{{\mathcal{S}}}

\def\sR{{\mathbb{R}}}

\newcommand{\E}{\ensuremath{\mathbb E}}           

\newcommand{\papertitle}{Feature-Based vs. GAN-Based Learning from Demonstrations: When and Why\xspace}

\usepackage{hyperref}
\usepackage{url}

\title{\papertitle}

\author{Chenhao Li \\
ETH AI Center \\
\texttt{chenhli@ethz.ch} \\
\And
Marco Hutter \\
ETH Zurich \\
\texttt{mahutter@ethz.ch} \\
\And
Andreas Krause \\
ETH Zurich \\
\texttt{krausea@ethz.ch} \\
}

\iclrfinalcopy 
\begin{document}

\maketitle

\begin{abstract}
This survey provides a comparative analysis of feature-based and GAN-based approaches to learning from demonstrations, with a focus on the structure of reward functions and their implications for policy learning.
Feature-based methods offer dense, interpretable rewards that excel at high-fidelity motion imitation, yet often require sophisticated representations of references and struggle with generalization in unstructured settings.
GAN-based methods, in contrast, use implicit, distributional supervision that enables scalability and adaptation flexibility, but are prone to training instability and coarse reward signals.
Recent advancements in both paradigms converge on the importance of structured motion representations, which enable smoother transitions, controllable synthesis, and improved task integration.
We argue that the dichotomy between feature-based and GAN-based methods is increasingly nuanced: rather than one paradigm dominating the other, the choice should be guided by task-specific priorities such as fidelity, diversity, interpretability, and adaptability.
This work outlines the algorithmic trade-offs and design considerations that underlie method selection, offering a framework for principled decision-making in learning from demonstrations.

\end{abstract}


\subsection*{Acknowledgments}
This survey was proofread by Zhiyang Dou, Tairan He, Xuxin Cheng, Zhengyi Luo, and Chen Tessler.
Their expertise in the field and constructive suggestions were instrumental in shaping the final form of this work.

\section{Disclaimer}
The terminology surrounding the use of offline reference data in reinforcement learning (RL) varies widely across the literature.
Terms such as \textit{imitation learning}, \textit{learning from demonstrations}, and \textit{demonstration learning} are often used interchangeably, despite referring to subtly different methodologies or assumptions.

In this survey, we adopt the term \textbf{learning from demonstrations} to specifically denote a class of methods that utilize \textit{state-based}, \textit{offline reference data} to derive a \textbf{reward signal}.
This reward signal quantifies the similarity between the behavior of a learning agent and that of the reference trajectories, and it is used to guide policy optimization.

This definition intentionally excludes methods based on \textbf{behavior cloning} that require \textbf{action annotations}, such as those used in recent large-scale manipulation datasets (e.g., Gr00t N1~\citep{bjorck2025gr00t}, diffusion policy~\citep{chi2023diffusion}, Gemini Robotics~\citep{team2025gemini}).
These approaches assume access to expert action labels and thus follow a different paradigm than the class of methods discussed here, which operate solely on state observations and rely on RL to generate control.

\section{Motivation and Scope}
While learning from demonstrations has become a widely adopted strategy in both robotics and character animation, the field lacks consistent guidance on \textbf{when to prefer particular classes of methods}, such as feature-based versus GAN-based approaches.
Practitioners often adopt one method over another based on precedent or anecdotal success, without a systematic analysis of the algorithmic factors that underlie their performance.
As a result, conclusions drawn from empirical success may conflate algorithmic merit with incidental choices in reward design, data selection, or architecture.

The objective of this article is to \textbf{provide a principled comparison between feature-based and GAN-based imitation methods}, focusing on their fundamental assumptions, inductive biases, and operational regimes.
The exposition proceeds in two stages.
First, we review the problem setting from the perspective of physics-based control and reinforcement learning, including the formulation of reward functions based on reference trajectories.
Second, we examine the historical development and current landscape of imitation methods, organized around the type of reward structure they use, explicit, feature-based formulations versus implicit, adversarially learned metrics.

Our goal is not to advocate for one approach over the other in general, but to clarify the conditions under which each is more suitable.
By articulating the trade-offs involved—including scalability, stability, generalization, and representation learning, we aim to provide a conceptual framework that supports more informed method selection in future work.

\section{Physics-Based Control, States and Actions}
In both character animation and robotics, \textbf{physics-based control} refers to a paradigm in which an agent's behavior is governed by the underlying physical dynamics of the system, either simulated or real.
Rather than prescribing trajectories explicitly, such as joint angles or end-effector poses, this approach formulates control as a process of goal-directed optimization, where a policy generates \textbf{control signals} (e.g., torques or muscle activations) to maximize an objective function under physical constraints.
This stands in contrast to \textbf{kinematics-based} or \textbf{keyframe-based} methods, which often disregard dynamics and focus on geometrically feasible but potentially physically implausible motions.
Physics-based control ensures that resulting behaviors are not only kinematically valid but also \textbf{dynamically consistent}, energy-conservative, and responsive to interaction forces, making it particularly suited for tasks involving locomotion, balance, and physical interaction in uncertain or dynamic environments.

The canonical formalism for this control paradigm is the \textbf{Markov Decision Process (MDP)}, defined by a tuple $\left (\gS, \gA, T, R, \gamma \right )$, where $\gS$ and $\gA$ denote the state and action spaces, respectively.
The transition kernel $T: \gS \times \gA \to \gS$ captures the environment dynamics $p \left (s_{t+1} \mid s_t, a_t \right )$, while the reward function $R: \gS \times \gA \times \gS \to \sR$ maps transitions to scalar rewards.
The agent seeks to learn a policy $\pi_\theta: \gS \to \gA$ that maximizes the expected discounted return $\E_{\pi_\theta} \left[\sum_{t \geq 0} \gamma^t r_t\right]$, where $r_t$ is the reward at time $t$ and $\gamma \in \left [0, 1 \right ]$ is the discount factor.

In this context, the state $s \in \gS$ typically encodes the agent's physical configuration and dynamics, such as joint positions, joint velocities, root orientation, and may include exteroceptive inputs like terrain geometry or object pose.
The action $a \in \gA$ corresponds to the control input applied to the system, most commonly joint torques in torque-controlled settings, or target positions in PD-controlled systems.
In biomechanical models, actions may also represent muscle activations.
By integrating these elements within a physics simulator or physical system, physics-based control enables emergent behaviors that are compatible with real-world dynamics, allowing policies to discover strategies that are not only effective but also physically feasible.

\section{Rethinking Learning from Demonstrations}

In the context of learning from demonstrations, reward functions are typically derived from \textbf{reference data}, rather than being manually engineered to reflect task success or motion quality.
This setup leverages recorded trajectories, often collected from motion capture, teleoperation, or other expert sources, to define a notion of behavioral similarity.
The policy is then optimized to minimize this discrepancy, encouraging it to reproduce motions that are consistent with those in the demonstration dataset.

Critically, the reward derived from demonstrations may serve either as a \textbf{pure imitation objective}, where the policy is expected to replicate the demonstrated behavior as closely as possible, or as a \textbf{regularizing component} that biases learning while allowing task-specific objectives to dominate.
This dual role makes demonstration-based rewards particularly valuable in high-dimensional control problems where exploration is difficult and task-based rewards are sparse or poorly shaped.
As such, learning from demonstrations transforms the design of the reward function from a manual engineering problem into one of defining or learning an appropriate similarity metric between agent and expert behavior, either explicitly, through features, or implicitly, through discriminators or encoders.

While reference trajectories are often valued for their visual realism or naturalness, this perspective underemphasizes their \textbf{algorithmic utility}: reference data serves as a critical mechanism for improving \textbf{learning efficiency} in high-dimensional control problems.
Rather than functioning merely as a constraint or prior, demonstrations provide \textbf{structured guidance} that biases policy exploration toward plausible and meaningful behaviors.

This role becomes especially important as the complexity of the environment and agent increases.
In lower-dimensional settings, carefully engineered reward functions or manually designed curricula have proven sufficient to elicit sophisticated behaviors through reinforcement learning alone~\citep{rudin2022learning}.
However, such strategies do not scale effectively to systems with high-dimensional state-action spaces, where naïve exploration is inefficient and reward shaping becomes brittle or intractable.
Under these conditions, demonstration data offers a \textbf{practical alternative to reward or environment shaping}, acting as an inductive bias that accelerates the discovery of viable behaviors.
In this light, reference motions are not ancillary constraints but \textbf{primary learning signals}, particularly in regimes where task-based supervision is sparse or difficult to specify.
This reframing justifies the use of demonstrations not only for imitation but as a foundation for scalable and data-efficient policy learning.

\section{Feature-Based Imitation: Origins and Limitations}
\begin{figure}
    \centering
    \includegraphics[width=0.9\linewidth]{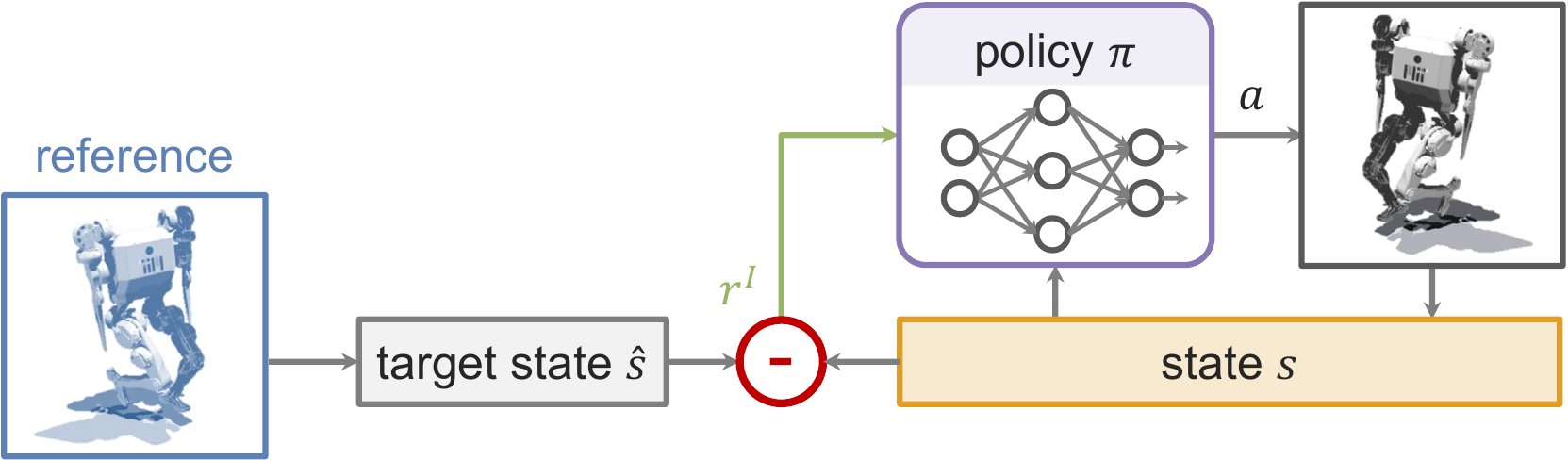}
    \caption{DeepMimic-style feature-based methods. The policy receives dense, per-frame rewards by comparing hand-crafted features—such as joint positions and end-effector poses—between its current state and a time-aligned reference state. A phase variable synchronizes policy and demonstration trajectories, enabling accurate motion reproduction but limiting generalization across diverse behaviors due to the lack of structured motion representation.}
    \label{fig:deepmimic}
\end{figure}

Feature-based imitation approaches can be traced back to DeepMimic~\citep{peng2018deepmimic}, which established a now-standard formulation for constructing reward signals based on explicit motion matching.
In this framework, the policy is aligned with a reference trajectory by introducing a \textbf{phase variable}, which serves as a learned proxy for temporal progress through the motion.
The reward is computed by evaluating feature-wise distances—such as joint positions, velocities, orientations, and end-effector positions—between the policy-generated trajectory and the reference, synchronized via the phase.
An abstracted overview is shown in \figref{fig:deepmimic}.

Owing to their dense and explicit reward structure, these methods are highly effective at reproducing \textbf{fine-grained motion details}.
However, their scalability to diverse motion datasets is limited.
While DeepMimic introduces a one-hot motion identifier to enable multi-clip training, this encoding does not model \textbf{semantic or structural relationships} between different motions.
As a result, the policy treats each motion clip as an isolated objective, which precludes generalization and often leads to discontinuities at transition points.

Although the phase variable handles temporal alignment within a given clip, there is no analogous mechanism for enforcing \textbf{spatial or semantic coherence} across clips.
Transitions between motions are implemented via hard switching on motion identifiers, which can result in abrupt behavioral changes and visually unnatural trajectories.
What is missing in this setup is a \textbf{structured representation space} over motions—one that captures both temporal progression and the underlying topology of behavioral variation.
Such representations enable not only smoother transitions between behaviors but also facilitate interpolation, compositionality, and improved generalization to motions not seen during training.
Policies trained over these structured motion spaces are better equipped to synthesize new behaviors while preserving physical plausibility and stylistic fidelity.

\section{Implicit Rewards for Motion Diversity: GAN-Based Imitation}
\begin{figure}
    \centering
    \includegraphics[width=0.9\linewidth]{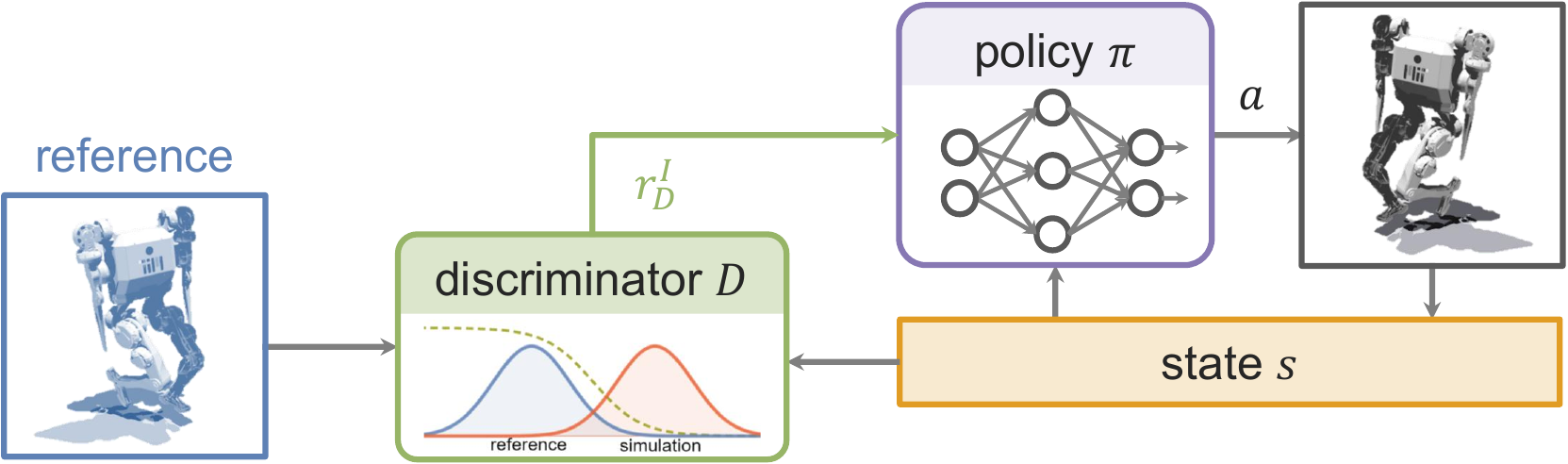}
    \caption{GAN-based methods via adversarial rewards. A discriminator learns to distinguish short transition snippets from policy-generated and demonstration data, providing an implicit reward signal that guides the policy toward expert-like behavior. By operating on short windows without explicit time alignment, this approach scales to diverse motion datasets and captures distributional similarity, enabling smoother transitions across unstructured behaviors.}
    \label{fig:gan-based}
\end{figure}

To address the limitations of feature-based approaches in handling diverse motion data, Adversarial Motion Priors (AMP)~\citep{peng2021amp} introduced the use of adversarial training, building on earlier frameworks such as GAIL~\citep{ho2016generative}, where expert action labels are assumed.
In the AMP setting, a \textbf{discriminator} is trained to distinguish between state transitions generated by the policy and those sampled from a dataset of reference trajectories.
As the policy improves, its transitions become increasingly indistinguishable from the expert data, thereby reducing the discriminator's ability to classify them correctly.
The discriminator’s output serves as a reward signal, guiding the policy toward behavioral fidelity.
The system is illustrated in \figref{fig:gan-based}.

From an optimization standpoint, GAN-based methods treat the policy as a \textbf{generator} in a two-player minimax game.
These methods scale naturally to large and diverse motion datasets, as they operate on short, fixed-length \textbf{transition windows}, typically spanning two to eight frames, rather than full trajectories.
This removes the need for phase-based or time-indexed alignment, making them particularly effective in unstructured datasets.
Additionally, the discriminator implicitly defines a \textbf{similarity metric} over motion fragments, allowing transitions that are behaviorally similar to receive comparable rewards even when not temporally aligned.
As a result, policies trained under adversarial objectives tend to exhibit smoother transitions across behaviors compared to methods relying on discrete motion identifiers and hard switching.
Because the reward is defined over \textbf{distributional similarity}, rather than matching a specific trajectory, AMP and related techniques are well-suited for stylization tasks or for serving as general motion priors that can be composed with task-specific objectives.

Despite their empirical success across domains, including character animation (e.g., InterPhys~\citep{hassan2023synthesizing}, PACER~\citep{rempe2023trace}) and robotics~\citep{escontrela2022adversarial}, adversarial imitation introduces fundamental challenges that impact training reliability and policy expressiveness.

\textbf{Discriminator Saturation} \; A key challenge in adversarial setups is that the discriminator can rapidly become overconfident, especially early in training when the policy generates trajectories that diverge significantly from the reference distribution.
In this regime, the discriminator easily classifies all transitions correctly, producing near-zero gradients and leaving the policy without informative reward signals.
This phenomenon is particularly problematic in high-dimensional or difficult environments, such as rough terrain locomotion or manipulation tasks, where meaningful exploration is essential but sparse.

Solutions such as Wasserstein-based objectives (e.g., WASABI~\citep{li2023learning}, HumanMimic~\citep{tang2024humanmimic}) aim to retain useful gradients and therefore reward signals even in the face of a strong discriminator.

\textbf{Mode Collapse} \; Another failure mode is the collapse of behavioral diversity: the policy may converge to producing only a narrow subset of trajectories that reliably fool the discriminator, ignoring the wider variation present in the demonstrations.
While the discriminator implicitly encourages local smoothness in the reward landscape, AMP lacks a structured motion representation that would enable global diversity or controllable behavior synthesis.
Consequently, the resulting policies often underutilize the full range of skills present in the data.

\begin{figure}
    \centering
    \includegraphics[width=0.9\linewidth]{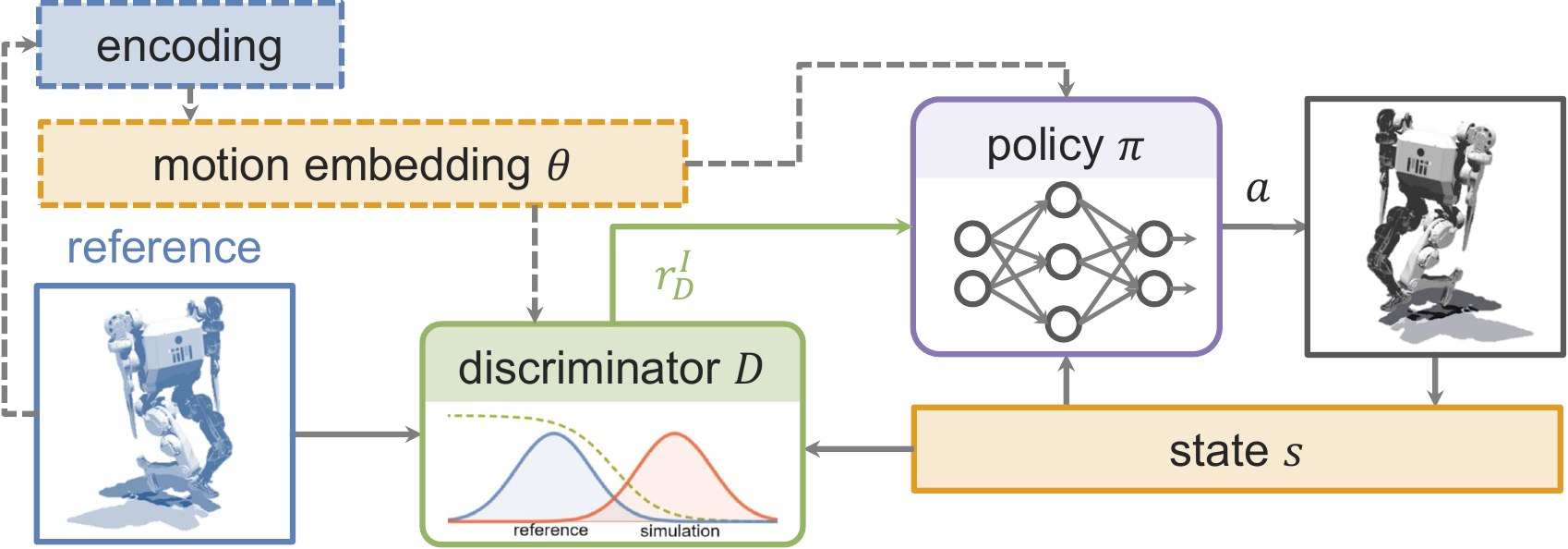}
    \caption{Latent-conditioned GAN-based methods. The policy and discriminator are jointly conditioned on learned motion embeddings, which are derived from demonstration data through unsupervised or supervised representation learning. These latent variables structure the imitation space, promoting behavioral diversity, stabilizing training, and enabling controllable skill generation beyond what implicit adversarial objectives can achieve alone.}
    \label{fig:gan-based_conditioned}
\end{figure}

To counteract this limitation, a variety of techniques introduce latent representations to provide structured control over motion variation as shown in \figref{fig:gan-based_conditioned}.

Unsupervised approaches like CASSI~\citep{li2023versatile}, ASE~\citep{peng2022ase}, and CALM~\citep{tessler2023calm} learn continuous embeddings over motion space, optimizing mutual information between latent codes and observed behaviors to preserve diversity.
These embeddings are then used to condition the policy, enabling the generation of distinct behaviors from different regions of the latent space.
Other approaches rely on category-level supervision to guide the learning process.
For example, Multi-AMP~\citep{vollenweider2023advanced}, CASE~\citep{dou2023c}, and SMPLOlympics~\citep{luo2024smplolympics} use motion class annotations to condition both the discriminator and the policy, thereby restricting collapse to occur only within class-specific subregions.
In contrast, FB-CPR~\citep{tirinzoni2025zero} adopts a representation-based solution, learning forward-backward encodings to structure the discriminator’s feedback.
Several other extensions train individual motion primitives progressively (e.g., PHC~\citep{luo2023perpetual}, PHC+~\citep{luo2023universal}).
A conditioned skill composer is utilized to recover the motion diversity.
Others introduce representation distillation with variational bottlenecks, as in PULSE~\citep{luo2023universal}, to form compressed yet expressive motion embeddings for controllable generation.

Together, these developments highlight both the flexibility and complexity of adversarial imitation learning.
While GAN-based methods naturally scale to large and diverse datasets, they benefit substantially from the addition of structured motion representations, whether learned, annotated, or composed, to stabilize training and recover controllable, diverse behavior.

\section{Feature-Based Imitation with Structured Representations}
\begin{figure}
    \centering
    \includegraphics[width=0.9\linewidth]{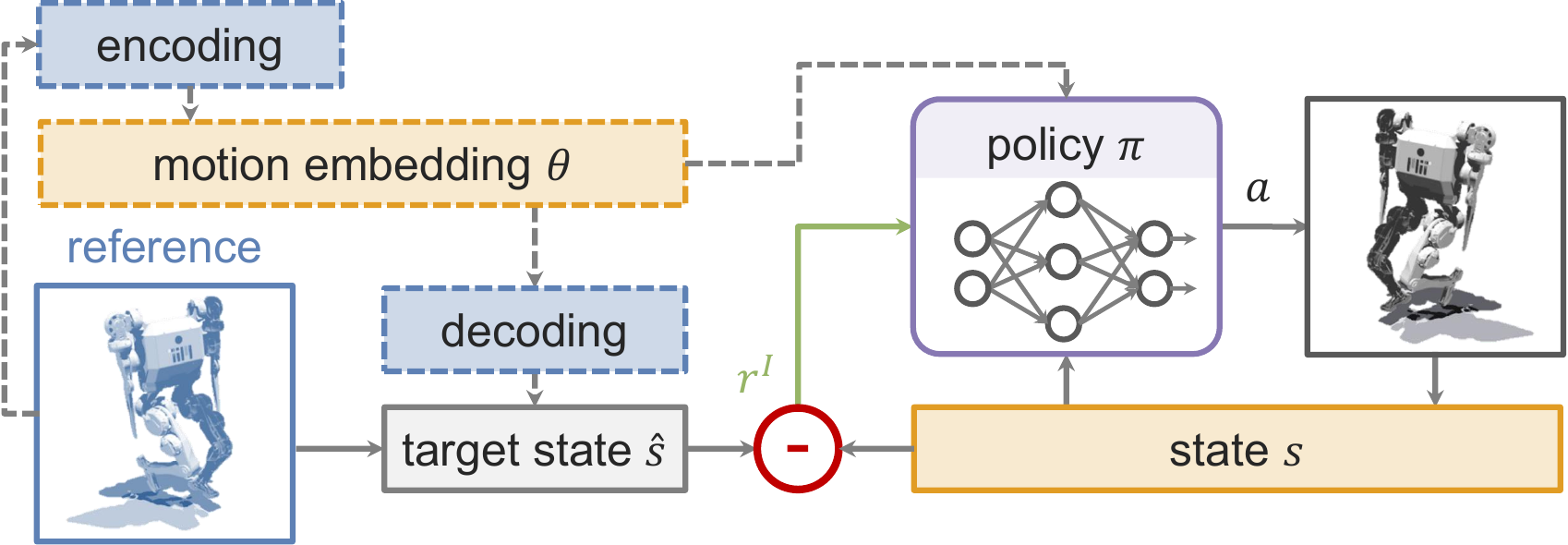}
    \caption{Feature-based methods with structured motion representations. The policy receives per-frame rewards based on feature differences with reference states and is conditioned on compact motion embeddings derived from demonstration data. This design preserves the interpretability of hand-crafted objectives while enabling smoother transitions and broader generalization across behaviors through learned motion structure.}
    \label{fig:feature-based}
\end{figure}

While adversarial imitation methods offer flexibility and scalability with diverse reference data, they impose significant practical burdens.
Ensuring training stability, managing discriminator saturation, and preventing mode collapse often require extensive architectural tuning.
These limitations have motivated a return to feature-based methods, now enhanced with structured motion representations, as a more interpretable and controllable alternative to adversarial training.
The core insight behind this renewed direction is the importance of a well-structured motion representation space for enabling smooth transitions and generalization across behaviors.
While GAN-based methods rely on the discriminator to \textbf{implicitly induce} such a representation, often requiring additional mechanisms to extract, control, or condition on it, feature-based approaches allow for the \textbf{explicit construction} of motion embeddings that are either precomputed or learned in parallel with policy training.
This explicitness simplifies conditioning and reward design, often reducing the reward to weighted feature differences relative to a reference state.
Such systems can be abstracted with a structure illustrated in \figref{fig:feature-based}.

As a result, a new class of imitation approaches has emerged that maintains the explicit reward structure of traditional feature-based methods, but augments it with \textbf{representation learning} to scale across tasks and motions.
In many cases, reference frames, or compact summaries thereof, are injected directly into the policy, providing frame-level tracking targets that guide behavior.

\textbf{Sophisticated Motion Representation} \; A central challenge for this class of methods is the construction of motion representations that support smooth transitions and structural generalization. Compact, low-dimensional embeddings promote semantic understanding of inter-motion relationships and improve sample efficiency.

To this end, some methods inject reference features or full motion states directly into the policy (e.g., PhysHOI~\citep{wang2023physhoi}, ExBody~\citep{cheng2024expressive}, H2O~\citep{he2024learning}, HumanPlus~\citep{fu2024humanplus}, MaskedMimic~\citep{tessler2024maskedmimic}, ExBody2~\citep{ji2024exbody2}, OmniH2O~\citep{he2024omnih2o}, AMO~\citep{li2025amo}, TWIST~\citep{ze2025twist}, GMT~\citep{chen2025gmt}), preserving spatial coherence in the motion space.
Others pursue more abstract embeddings through self-supervised or policy-conditioned learning.
For instance, ControlVAE~\citep{yao2022controlvae}, PhysicsVAE~\citep{won2022physics}, and NCP~\citep{zhu2023neural} build representations via policy interaction, while VMP~\citep{serifi2024vmp} and RobotMDM~\citep{serifi2024robot} construct temporally and spatially coherent embeddings using self-supervision.
Frequency-domain methods such as PAE~\citep{starke2022deepphase}, FLD~\citep{li2024fld}, and DFM~\citep{watanabe2025dfm} impose motion-inductive biases that capture the periodic and hierarchical structure of motion.
These techniques collectively extend the DeepMimic paradigm by generalizing phase alignment and structural similarity beyond heuristics.

\textbf{Inflexible Imitation Adaptation} \; A limitation of these representation-driven feature-based methods is that they often rely on explicit tracking of full trajectories, enforced by dense per-step rewards.
This design makes it difficult to adapt or deviate from the reference when auxiliary tasks require flexibility, as is common in goal-directed or interaction-heavy settings.

To address this, some approaches introduce mechanisms to adaptively relax imitation constraints.
For example, MCP~\citep{sleiman2024guided} introduces a fallback mechanism that adjusts phase progression when key task objectives are not met.
RobotKeyframing~\citep{zargarbashi2024robotkeyframing} proposes a transformer-based attention model that encodes arbitrary sets of keyframes with flexible temporal spacing.
ConsMimic~\citep{wen2025constrained} proceeds with imitation of features only when the optimality constraints of the task are satisfied.
Other works incorporate high-level planning components to dictate intermediate reference states, such as diffusion-based models in PARC~\citep{xu2025parc} and HMI~\citep{fan2025one}, or planners that directly modulate the learned motion representations (e.g., VQ-PMC~\citep{han2024lifelike}, Motion Priors Reimagined~\citep{zhang2025motion}).

Together, these developments illustrate the interpretability and stability of feature-based imitation when paired with structured motion representations.
However, despite avoiding the instability of adversarial training, these methods remain constrained by their reliance on explicit tracking and overengineered representations, which can hinder adaptation in tasks requiring flexible deviation from demonstrations.

\section{Summary: Strengths, Limitations, and Emerging Directions}
Learning from demonstrations has evolved into two primary methodological paradigms: \textbf{feature-based methods}, which use explicit, hand-crafted reward formulations, and \textbf{GAN-based methods}, which employ discriminators to implicitly shape behavior.
Each offers distinct advantages and faces unique challenges, especially as the field shifts toward learning from large, diverse, and unstructured motion datasets.
We summarize the aforementioned works in \tabref{table:taxonomy}.

\begin{table}[h]
    \centering
    \caption{Taxonomy of learning from demonstration methods.}
    \begin{tabular}{l}
    \toprule
    \textbf{GAN-based} \\
    \midrule
    AMP~\citep{peng2021amp, escontrela2022adversarial}, InterPhys~\citep{hassan2023synthesizing}, \\
    PACER~\citep{rempe2023trace}, WASABI~\citep{li2023learning}, HumanMimic~\citep{tang2024humanmimic}, \\
    CASSI~\citep{li2023versatile}, ASE~\citep{peng2022ase}, CALM~\citep{tessler2023calm}, \\
    Multi-AMP~\citep{vollenweider2023advanced}, CASE~\citep{dou2023c}, \\
    SMPLOlympics~\citep{luo2024smplolympics}, FB-CPR~\citep{tirinzoni2025zero}, PHC~\citep{luo2023perpetual}, \\ PHC+~\citep{luo2023universal}, PULSE~\citep{luo2023universal} \\
    \midrule
    \textbf{Feature-based} \\
    \midrule
    DeepMimic~\citep{peng2018deepmimic}, PhysHOI~\citep{wang2023physhoi}, ExBody~\citep{cheng2024expressive}, \\
    H2O~\citep{he2024learning}, HumanPlus~\citep{fu2024humanplus}, MaskedMimic~\citep{tessler2024maskedmimic}, \\
    ExBody2~\citep{ji2024exbody2}, OmniH2O~\citep{he2024omnih2o}, AMO~\citep{li2025amo}, \\
    TWIST~\citep{ze2025twist}, GMT~\citep{chen2025gmt}, ControlVAE~\citep{yao2022controlvae}, \\
    PhysicsVAE~\citep{won2022physics}, NCP~\citep{zhu2023neural}, VMP~\citep{serifi2024vmp}, \\
    RobotMDM~\citep{serifi2024robot}, PAE~\citep{starke2022deepphase}, FLD~\citep{li2024fld}, \\
    DFM~\citep{watanabe2025dfm}, MCP~\citep{sleiman2024guided}, ConsMimic~\citep{wen2025constrained}, \\
    RobotKeyframing~\citep{zargarbashi2024robotkeyframing}, PARC~\citep{xu2025parc}, HMI~\citep{fan2025one}, \\
    VQ-PMC~\citep{han2024lifelike}, Motion Priors Reimagined~\citep{zhang2025motion} \\
    \bottomrule
    \end{tabular}
    \label{table:taxonomy}
\end{table}

\subsection{GAN-Based Methods}

GAN-based approaches, such as AMP and its derivatives, use a discriminator to assign reward signals based on the realism of short transition snippets.
This formulation dispenses with time-aligned supervision, allowing policies to imitate motion in a distributional sense rather than reproducing specific trajectories.
As a result, these methods scale naturally to unstructured or unlabeled data, enabling smoother transitions between behaviors and generalization beyond the demonstrated clips.

Recent advances mitigate some of the core challenges of GAN-based imitation, namely, \textbf{discriminator saturation} and \textbf{mode collapse}, by introducing latent structure.
Techniques learn motion embeddings that condition both policy and discriminator, thereby stabilizing training and supporting controllable behavior generation.
These latent-conditioned GANs can also model semantic structure in motion space, facilitating interpolation and compositionality.

Despite these benefits, GAN-based methods remain prone to \textbf{training instability}, require careful discriminator design, and often offer coarser control over motion details.
Their implicit reward structure can obscure performance tuning and requires auxiliary mechanisms for precise task alignment.

\subsection{Feature-Based Methods}

In contrast, feature-based imitation methods like DeepMimic start with dense, per-frame reward functions derived from specific motion features.
This yields strong supervision for motion matching, making them highly effective for replicating fine-grained details in demonstrated behavior.
However, traditional approaches are limited by their \textbf{dependence on hard-coded alignment} and \textbf{lack of structured motion representation}, which restricts scalability and generalization.

Recent developments address these limitations by integrating learned motion representations into the reward and policy structure. 
These efforts construct latent motion embeddings to structure behavior across clips, enabling smoother transitions and support for more diverse or compositional motions.
This new generation of feature-based methods retains interpretability and strong reward signals while gaining some of the flexibility previously unique to GAN-based setups.

Nevertheless, feature-based systems still face challenges in \textbf{adapting to auxiliary tasks or goals} that require deviation from the reference trajectory.
Their strong reliance on explicit tracking and dense supervision can make them brittle in dynamic or multi-objective settings, where flexibility is crucial.

\begin{table}[h]
    \centering
    \caption{Comparative analysis.}
    \begin{tabular}{lll}
    \toprule
    \textbf{Criterion} & \textbf{GAN-Based Methods} & \textbf{Feature-Based Methods} \\
    \midrule
    Reward signal & implicit, coarse & explicit, dense \\
    \midrule
    Scalability & high (unstructured data) & moderate (depends on representation) \\
    \midrule
    Generalization & strong with latent conditioning & strong with good embeddings \\
    \midrule
    Training stability & challenging (saturation, collapse) & stable but sensitive to inductive bias \\
    \midrule
    Interpretability & low to moderate & high \\
    \midrule
    Control & indirect (via discriminator or latent) & direct (via features or embeddings) \\
    \midrule
    Task integration & flexible & precise but less adaptable \\
    \bottomrule
    \end{tabular}
    \label{table:comparative_analysis}
\end{table}

\section{On Metrics and Misconceptions}
In evaluating learning from demonstration algorithms, it is common practice to reference metrics such as \textit{motion naturalness}, \textit{energy efficiency}, or \textit{cost of transport}.
While these properties are intuitively appealing, they can be misleading indicators of algorithmic performance.
Crucially, such metrics are not inherent to the learning algorithm itself but are instead highly dependent on the quality and structure of the reference data.
For instance, if a policy trained via a particular algorithm exhibits smoother or more energy-efficient behavior, this outcome often reflects characteristics of the underlying demonstrations rather than advantages intrinsic to the algorithmic formulation.
Consequently, attributing these observed properties to the learning method risks conflating algorithmic capability with dataset bias.

Moreover, these high-level metrics offer limited diagnostic value when comparing algorithm classes.
They do not capture fundamental differences in reward design, training stability, scalability, or generalization capacity.
A GAN-based approach may yield visually smoother transitions due to its distributional objectives, but this benefit must be weighed against the challenges of motion diversity and tracking accuracy.
Conversely, a feature-based method may produce high-fidelity imitation in terms of kinematic features but struggle with generalization due to its reliance on well-structured representations.
To conduct a rigorous and meaningful comparison between methods, evaluation should focus on the properties most directly influenced by algorithmic design.
These include reward signal quality, training stability, generalization to novel motions or environments, and adaptability to auxiliary tasks.
By focusing on such factors, researchers and practitioners can better understand the operational trade-offs between feature-based and GAN-based approaches, avoiding overgeneralized claims and grounding comparisons in algorithmic substance rather than incidental outcome metrics.

\section{Debunking Common Beliefs}
Despite a growing body of research, misconceptions remain prevalent in discussions of GAN-based versus feature-based learning from demonstrations.
Below, we revisit some common claims, clarify their limitations, and situate them within a more rigorous analytical framework.

\begin{quote}
    \textbf{``GAN-based methods automatically develop a distance metric between reference and policy motions.''}
\end{quote}

This is partially true.
GAN-based methods implicitly learn a similarity function via the discriminator.
However, this function may be ill-defined in early training, leading to discriminator saturation, where the discriminator assigns uniformly high distances regardless of policy improvement.
Moreover, the discriminator may conflate resemblance to a single exemplar with similarity to the overall distribution, resulting in mode collapse.
Thus, while a learned metric exists, its utility and stability depend heavily on discriminator design and representation quality.

\begin{quote}
    \textbf{``GAN-based methods do not require hand-crafted features.''}
\end{quote}

No.
This assertion overlooks a key implementation detail: the discriminator operates on selected features of the agent state.
Choosing these features is analogous to defining reward components in feature-based methods.
Insufficient features can prevent the discriminator from detecting meaningful discrepancies, while overly complex inputs can lead to rapid overfitting and saturation.
This trade-off is particularly critical in tasks involving partially observed context (e.g., terrain or object interactions), where feature selection significantly impacts training stability and convergence.

\begin{quote}
    \textbf{``GAN-based methods avoid hand-tuned reward weights for different features.''}
\end{quote}

Not quite.
While adversarial methods circumvent explicit manual weighting of reward components, they are still sensitive to feature scaling and normalization.
Input magnitudes shape the discriminator’s sensitivity and therefore act as an implicit weighting scheme.
Poorly calibrated inputs can bias the reward signal, undermining the interpretability and reliability of the learned policy.

\begin{quote}
    \textbf{``GAN-based methods yield smoother transitions between motions.''}
\end{quote}

This holds true only relative to early feature-based methods that lacked structured representations and relied on hard switching between clips.
Modern feature-based methods that leverage structured motion embeddings can produce smooth, semantically meaningful transitions.
Interpolation in learned latent spaces supports temporally and spatially coherent motion generation, rivaling or exceeding GAN-based transitions when appropriate representation learning is applied.

\begin{quote}
    \textbf{``Only GAN-based methods can be combined with task rewards.''}
\end{quote}

No.
Both GAN-based and feature-based methods can incorporate task objectives.
Feature-based methods provide dense, frame-aligned imitation rewards, making them effective when the task aligns closely with the reference motion, but less flexible when deviation is required.
In contrast, GAN-based methods offer distribution-level supervision, enabling greater adaptability to auxiliary goals.
This flexibility, however, comes at the cost of lower fidelity to the reference and a risk of mode collapse.

\begin{quote}
    \textbf{``GAN-based methods deal better with unstructured or noisy reference motions.''}
\end{quote}

This is an oversimplification.
GAN-based methods can exhibit robustness to small inconsistencies in demonstrations due to their distributional supervision.
However, this robustness often comes at the cost of discarding fine motion details.
Feature-based approaches, especially those employing probabilistic or variational models, can also handle noise effectively through regularization and representation smoothing.

\begin{quote}
    \textbf{``GAN-based methods scale better.''}
\end{quote}

Not necessarily.
Scalability is more a function of motion representation quality than of paradigm.
Both GAN-based and feature-based methods can scale with large datasets if equipped with appropriate latent encodings.
The difference lies in when and how these representations are learned—feature-based methods often rely on supervised or self-supervised embeddings, while GAN-based methods may induce representations via adversarial feedback.
Neither approach guarantees scalability without careful design.

\begin{quote}
    \textbf{``GAN-based methods transfer better to real-world deployment.''}
\end{quote}

No.
There is no intrinsic connection between the choice of imitation algorithm and sim-to-real transfer efficacy.
Transferability is determined primarily by external strategies such as domain randomization, system identification, and regularization.
While GAN-based approaches may respond more flexibly to auxiliary rewards, they are also more sensitive to regularization, which can create the false impression that certain regularizers are more effective in these methods.

\begin{quote}
    \textbf{``Feature-based methods generalize better to unseen motion inputs.''}
\end{quote}

Generalization depends less on the reward structure and more on the quality and organization of the motion representation space.
Both GAN-based and feature-based methods can generalize effectively when equipped with well-structured embeddings.
Failure modes arise not from the paradigm itself but from inadequate inductive biases, insufficient diversity in training data, or poor temporal modeling.

\begin{quote}
    \textbf{``Feature-based methods are easier to implement.''}
\end{quote}

Not necessarily.
Designing robust feature-based systems involves selecting appropriate reward features, constructing phase functions or embeddings, and managing temporal alignment.
These tasks can be as complex as designing a discriminator, particularly when the goal is to scale across tasks or environments.
Moreover, effective latent representations often require pretraining and careful architectural choices to avoid collapse or disentanglement failure.

\section{Final Remarks}
This survey has examined two major paradigms in learning from demonstrations: feature-based and GAN-based methods, through the lens of reward structure, scalability, generalization, and representation.
The core distinction lies not merely in architectural components but in their respective philosophies of supervision: explicit, hand-crafted rewards versus implicit, adversarially learned objectives.

\textbf{Feature-based methods} offer dense, interpretable rewards that strongly anchor the policy to reference trajectories, making them well-suited for tasks requiring high-fidelity reproduction of demonstrated motions.
However, they often struggle with generalization, particularly in multi-clip or unstructured settings, due to the need for manually specified features and aligned references.

\textbf{GAN-based methods}, in contrast, provide more flexible and data-driven reward structures through discriminative objectives.
This enables them to scale naturally to diverse datasets and to support smoother transitions and behavior interpolation.
Yet, they often encounter challenges related to training stability, reward sparsity, and loss of fine-grained motion detail.

It is important to recognize that \textbf{many problems commonly attributed to one paradigm reappear in different forms in the other}.
For instance, mode collapse in GANs mirrors the brittleness of poor motion representations in feature-based methods.
Similarly, while feature-based methods offer strong guidance for motion tracking, they may fail to generalize or adapt when rigid reward definitions are misaligned with auxiliary tasks or dynamic environments.

Rather than presenting these two paradigms as mutually exclusive, recent trends point toward a \textbf{convergent perspective}, one that emphasizes the centrality of structured motion representations.
Whether derived from self-supervised learning, latent encodings, or manually designed summaries, these representations serve as a bridge between the strengths of each approach: the interpretability and controllability of explicit rewards and the scalability and adaptability of adversarial training.

Ultimately, the decision between using a feature-based or GAN-based approach is not a question of universal superiority.
Instead, it should be guided by the \textbf{specific constraints and priorities of the application}: fidelity versus diversity, interpretability versus flexibility, or training simplicity versus large-scale generalization.
Understanding these trade-offs and their relationship to reward structure and motion representation is essential for designing robust, scalable, and expressive imitation learning systems.

\clearpage

\bibliography{main}
\bibliographystyle{iclr2024_conference}



\end{document}